\documentclass[letterpaper, 10 pt, conference]{ieeeconf}  %

\IEEEoverridecommandlockouts                              %

\usepackage{times}
\usepackage{epsfig}
\usepackage{graphicx}
\usepackage{amsmath}
\usepackage{amssymb}
\usepackage{xspace}
\usepackage{booktabs}

\usepackage{amsfonts}

\usepackage{epstopdf}
\usepackage{color}

\usepackage{multirow}
\usepackage{pbox}

\usepackage{amsopn}

\usepackage{multirow}
\usepackage{longtable}
\usepackage{array}
\usepackage{makecell}

\usepackage{cellspace}
\makeatletter

\DeclareRobustCommand\onedot{\futurelet\@let@token\@onedot}
\def\@onedot{\ifx\@let@token.\else.\null\fi\xspace}

\def\eg{\emph{e.g}\onedot} 
\def\ie{\emph{i.e}\onedot} 
 
\def\etc{\emph{etc}\onedot} 
 
\def\etal{\emph{et al}\onedot}

\newcommand{\PAR}[1]{\vskip4pt \noindent {\bf #1~}}
\newcommand{\PARbegin}[1]{\noindent {\bf #1~}}

\newcommand{\categoryname}[1]{\textit{#1}}

\title{\LARGE \bf
Large-Scale Object Mining for Object Discovery from Unlabeled Video
}

\author{Aljo\v{s}a O\v{s}ep*, Paul Voigtlaender*, Jonathon Luiten, Stefan Breuers, Bastian Leibe%
\thanks{
\scriptsize \textsuperscript{*} Equal contribution.
\scriptsize The authors are with the Visual Computing Institute, RWTH Aachen University. %
E-mail: {\tt\scriptsize lastname@vision.rwth-aachen.de}}%
}

\begin{document}

\maketitle
\thispagestyle{empty}
\pagestyle{empty}

\begin{abstract}

This paper addresses the problem of object discovery from unlabeled driving videos captured in a realistic automotive setting. Identifying recurring object categories in such raw video streams is a very challenging problem. Not only do object candidates first have to be localized in the input images, but many interesting object categories occur relatively infrequently. Object discovery will therefore have to deal with the difficulties of operating in the long tail of the object distribution.
We demonstrate the feasibility of performing fully automatic object discovery in such a setting by mining object tracks using a generic object tracker. In order to facilitate further research in object discovery, we release a collection of more than $360,\!000$ automatically mined object tracks from $10+$ hours of video data ($560,\!000$ frames).
We use this dataset to evaluate the suitability of different feature representations and clustering strategies for object discovery.

\end{abstract}

\section{Introduction}

Deep learning has revolutionized the way research is being performed in computer vision, and the success of this development holds great promise for important applications such as autonomous driving~\cite{Janai17ARXIV}.
However, deep learning requires huge quantities of annotated training data, which are very costly to obtain. Consequently, progress has so far been limited to areas where such data is available, and community efforts such as PASCAL VOC~\cite{Everingham10IJCV}, ImageNet~\cite{Deng09CVPR}, CalTech~\cite{Dollar12PAMI}, KITTI~\cite{Geiger12CVPR}, COCO~\cite{Lin14ECCV}, or Cityscapes~\cite{Cordts16CVPR} have been instrumental in enabling recent successes. It is largely thanks to those efforts that we nowadays have good object detectors (\eg,~\cite{Ren15NIPS,Redmon16CVPR,Liu16ECCV}) at our disposal for a limited number of 20-80 object categories.

\begin{figure}[t]
	\begin{center}
   		\includegraphics[width=1.0\linewidth]{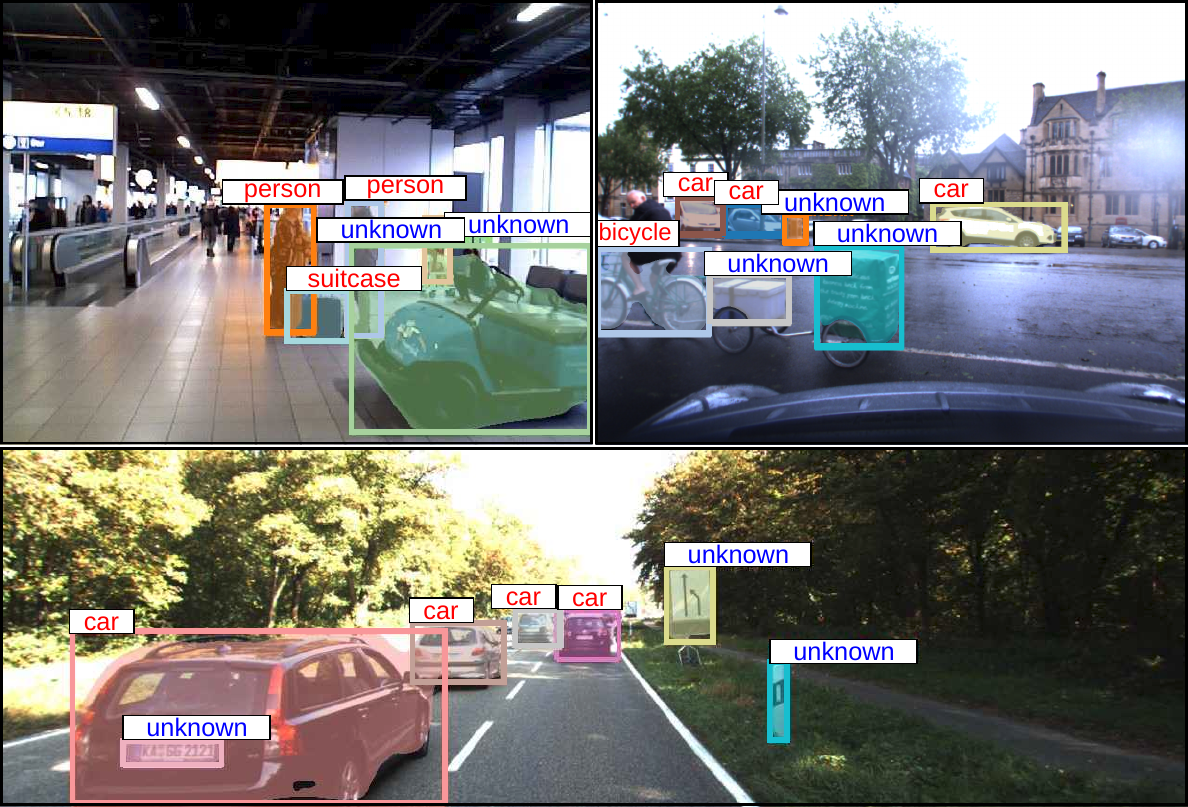}
	\end{center}
	\vspace{-15pt}
   \caption{We propose an approach for automatic discovery of novel and rare object categories from large video corpora. We start by mining generic object tracks (see above) and extract novel object categories by applying clustering using unknown object tracks.}
\label{fig:long}
\label{fig:covergirl}
\end{figure}

When moving from image interpretation tasks to video understanding problems, however, it becomes clear that the current strategy of using exhaustive human annotation will quickly become infeasible. This problem is of particular relevance in autonomous driving and mobile robotics, where future intelligent agents will have to deal with a large variety of driving scenarios involving a multitude of relevant object classes, many of which are not captured by today's detectors (see Fig.~\ref{fig:covergirl}). 
In this paper, we explore an automatic approach for discovering novel object categories (\ie, categories for which we do not have detectors yet) by mining generic object tracks from large driving video collections.

Object category discovery has attracted a lot of attention from the research community recently, and many approaches have been proposed for discovering object categories from image collections~\cite{Russell06CVPR,Rubinstein13CVPR,Tuytelaars10IJCV,Hsu16ARXIV,Song16CVPR,Hickinson17NIPSW} or videos~\cite{Kwak15ICCV,Tsai16ECCV}.
However, many current discovery approaches are evaluated on carefully pre-processed datasets, such as MNIST, CIFAR, or subsets of ImageNet~\cite{Hsu16ARXIV,Song16CVPR}, where each image contains an object of interest and the number of categories is a-priori known. We argue that such a setting is very different from real-world scenarios, where a major aspect of the difficulty of object discovery will be to deal with the long tail of the object distribution. In any practical application scenario, we can expect the frequency of object category observations to follow a power law distribution, with some object categories occurring very frequently and the vast majority being increasingly rare. Thus, even if every training example shows a potential object of interest, many rare object categories will not accumulate enough instances to allow clustering approaches to easily pick them out from the background noise. In order to make progress on this important topic, it therefore becomes important to focus evaluation on more realistic settings.

At the core of any object discovery approach is the question what constitutes an object. A common approach is to define object regions by a consistent appearance that separates the region from the surrounding background~\cite{Alexe12TPAMI}. In the literature, this definition has been adopted by region proposal networks~\cite{Ren15NIPS} that have been used successfully for object detection~\cite{Ren15NIPS,He17ICCV} and object discovery in internet images~\cite{Rubinstein13CVPR}. However, in our experience, such region proposals are not stable and not distinctive enough to permit generic object tracking in real street scenes.
We therefore adopt a farther-reaching definition of generic objects as \textit{regions that have well defined and temporally consistent boundaries in 3D space}. %
Further, as \categoryname{known} objects, we consider those for which we have a pre-trained detector available (\ie, the 80 annotated object categories in COCO~\cite{Lin14ECCV}), all the rest we consider to be \categoryname{unknown} objects.

In summary, we present a large-scale study for object mining and category discovery on two large datasets (KITTI Raw~\cite{Geiger13IJRR} and Oxford RobotCar~\cite{Maddern17IJRR}) for autonomous driving, comprising altogether roughly 10 hours of video data consisting of more than $560,\!000$ frames. From this data, we extract  more than $360,\!000$ object tracks using a fully automatic generic object tracking pipeline. As verified in our experiments, although the object tracks are extracted without human supervision and comprise both \categoryname{known} and \categoryname{unknown} object categories, less than 10\% of them are affected by tracking errors. Thus, they can serve as a stable basis for object discovery experiments. We use this dataset to evaluate the suitability of different feature representations and clustering strategies for object discovery.
To the best of our knowledge, this is the first time such a large-scale generic object mining effort has been undertaken in an automotive scenario. We make our code, datasets, and annotations publicly available to serve as a benchmark for the research community\footnote{\scriptsize Project website: \tt https://vision.rwth-aachen.de/page/lsom}.

\section{Related Work}

\PARbegin{Object Discovery.} Object discovery denotes the problem of identifying previously unseen object categories without human supervision.
Russell \etal~\cite{Russell06CVPR} propose a vision-based method that uses multiple object segmentations in order to group visually similar objects and their segmentations.
Sivic \etal~\cite{Sivic08CVPR} propose a method for discovery of hierarchical structure of objects from unlabeled images.
Lee and Grauman~\cite{Lee11CVPR} propose an iterative procedure that starts with easy-to-discover instances and progressively expands to more challenging cases and demonstrates that recognition in the form of a region classifier helps with the discovery by narrowing down object candidates~\cite{Lee10CVPR}.
We similarly utilize multiple object hypotheses in the form of tracklet proposals and utilize a classifier that assigns semantic information to tracklets.
Rubinstein \etal~\cite{Rubinstein13CVPR}
propose to identify potential objects in Internet images using saliency to find reoccurring patterns between images using dense correspondences. For a more detailed overview of existing image-based methods we refer to~\cite{Tuytelaars10IJCV}.
Kwak \etal~\cite{Kwak15ICCV} propose a method for joint tracking and object discovery in videos. Their method localizes and tracks the dominant object based on motion and saliency cues in each video. A similar idea is applied in Tsai \etal~\cite{Tsai16ECCV} for semantic co-segmentation in videos.
Both methods demonstrate excellent results on the YouTube-Objects dataset~\cite{Prest12CVPR}. However, these video sequences are usually dominated by a single object and cover only a limited number (10) of categories.

In the field of mobile robotics,~\cite{Endres09RSS, Triebel10RSS, Herbst11IROS} propose methods in which RGB-D scans are segmented into object candidates. These candidates are then grouped using either clustering methods or %
based on probabilistic inference.
However, all of these methods were only applied to simple indoor scenarios, containing well-separated objects such as boxes and chairs.
In~\cite{Zhang13ICRA, Moosmann11ICCVW} object discovery in traffic scenarios using LiDAR sensors is addressed. %
While for clean LiDAR data even simple methods can be used to segment scans into meaningful regions, obtaining object candidates from image data is far more challenging~\cite{Hosang14BMVC,Zitnick14ECCV, Alexe12TPAMI, Chen15NIPS, Osep16ICRA}.

\PAR{Clustering and Embedding Learning.}
Clustering is typically used to find patterns in unlabeled data by grouping data points by their similarity. Here the main challenge is defining similarities or distance measures between the data points. Recent methods approach this problem by learning distance metrics~\cite{Xie16ICML, Schroff15CVPR, Song17CVPR}. In order to adapt learned embeddings to a specific domain,~\cite{Yang16CVPR} proposes to iteratively cluster data and re-learn embeddings.
Hsu \etal~\cite{Hsu16ARXIV} propose a method that %
use a separate, labeled dataset to learn a Similarity Prediction Network (SPN),
which is then used to produce binary labels for each pair of objects of an unlabeled dataset. These labels are
used to train ClusterNet, which directly predicts cluster labels.
In contrast to the above-mentioned methods, we work with raw image data, where object localization is not given and we do not make any assumptions about object categories.

\PAR{Video-Object Mining.} Video-Object mining (VOM) refers to a task of collecting frequently-occurring patterns (\ie object candidates) from video or streams of sensory recordings in general.
Teichman \etal~\cite{Teichman12IJRR} propose a method for tracking-based semi-supervised learning by mining LiDAR streams, captured from a vehicle.
Similarly,~\cite{Misra15CVPR, Misra16ECCV} propose tracking-based semi-supervised learning based on video. 
Furthermore, in the context of vision, VOM has been used for improving object detectors by mining hard-negatives for specific object categories from web-videos~\cite{Tang12NIPS, Jin18ECCV}
and for learning new detectors for objects by localizing dominant video tubes in YouTube videos~\cite{Prest12CVPR}.

\section{Method}
\begin{figure*}[t]
\begin{center}
\includegraphics[width=1.0\linewidth]{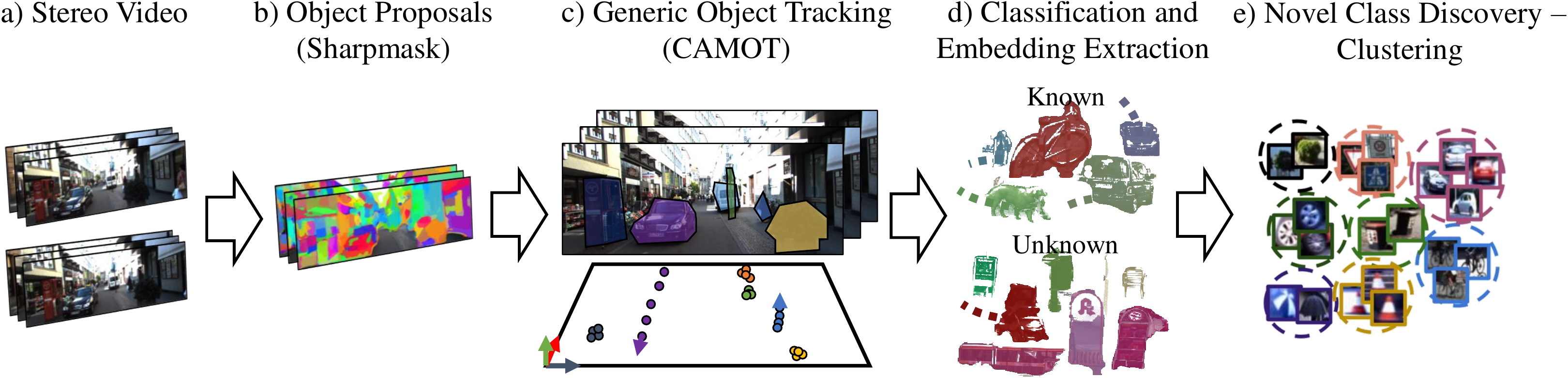}
\end{center}
\vspace{-15pt}
\caption{Our proposed method processes large amounts of stereo video data (a) using a generic object tracker (b-c). We compute track embedding vectors (d) that allow us to perform clustering efficiently in Euclidean space using standard clustering approaches, such as (H)DBSCAN or KMeans (e). This way, we can discover novel object categories among previously unknown (non-recognized) object tracks.}
\label{fig:pipeline}
\end{figure*}
Towards the goal of object discovery %
using unlabeled video sequences, we first need to be able to obtain potential object candidates in these video streams. There are large amounts of unlabeled video data available~\cite{Maddern17IJRR}, but finding new patterns in such data is challenging, as
state-of-the-art object proposal methods such as Sharpmask~\cite{Pinheiro16ECCV} need to produce 100-1000 proposals per frame to achieve a high recall. This would result in a very large set of object candidates on the level of an entire video.

We propose to leverage temporal information and prior knowledge about common object categories. By forming object tracks from image-level object proposals (Fig.~\ref{fig:pipeline}), we i) reduce the object candidate space considerably and ii) suppress noise and clutter in image-level object proposals, as these are typically unable to form stable object tracks. Recognition of common object types additionally helps reducing the proposal space and helps to suppress noise (see Tab.~\ref{tab:tracks-dataset}). These object candidates form our track collection.

\subsection{Object Track Mining}
\label{subsec:obj-track-mining}

For tracking, we build upon our recent work and utilize our category-agnostic multi-object tracker (CAMOT)~\cite{Osep18ICRA}.
In a nutshell, using this tracker object candidates are obtained as follows (see Fig.~\ref{fig:pipeline}).
The tracker takes as input stereo images and frame-level mask proposals from Sharpmask~\cite{Pinheiro16ECCV}.
CAMOT then uses these proposals in order to create a set of category-agnostic tracks. Afterwards, the classifier component of a Faster R-CNN~\cite{Ren15NIPS} based detector (trained on the COCO dataset) is used to classify the tracks on an image crop level.

Tracks are thus automatically labeled by the recognized category type (\ie, as one of the COCO~\cite{Lin14ECCV} categories) or as \categoryname{unknown} object track. 
Finally, for each frame, a mutually consistent subset of tracks is picked by performing MAP inference using a conditional random field (CRF) model (for details, see~\cite{Osep18ICRA}). 
This way, we obtain a reduced set of object tracks.
Tracks that are labeled as \categoryname{unknown} are considered object candidates and are used for object discovery.

\PAR{Track Postprocessing.} After applying the tracker, we obtain a large collection of selected tracklets of both \categoryname{known} and \categoryname{unknown} categories. Since model selection is performed on a per-frame basis, one object may be split into several short tracklets. As a final postprocessing step extending our tracker~\cite{Osep18ICRA} we progressively merge short selected tracklets into final tracks. In each frame, either a) existing tracklet $h_i$ is re-selected and trivially continues an existing track $H_k$, or b) tracklet $h_i$ is not re-selected and its track is continued by another selected tracklet $h_j$ if they have a sufficient overlap. If $h_i$ and $h_j$ do not have sufficient overlap, $H_k$ is terminated and $h_j$ starts a new track. As an overlap criterion, we use the fraction of matching masks to the length of the shorter tracklet:
\begin{equation}
\lambda \left ( h_i, h_j \right ) = \frac{\left | \left \{  t  |  IoU \left ( h_i^t, h_j^t \right ) > \gamma  \right \} \right |}{\text{min} \left ( \left | h_i \right |, \left | h_j \right |  \right )}.
\end{equation}
Here, two masks are considered to be a match when mask IoU is higher than a threshold $\gamma$ in frame $t$.

\PAR{Video Mining.} We applied the tracker on two publicly available datasets, KITTI Raw~\cite{Geiger13IJRR} and Oxford RobotCar~\cite{Maddern17IJRR}. 
For both we use stereo for estimating depth~\cite{Geiger10ACCV}. 
Compared to the original CAMOT~\cite{Osep18ICRA},
we replace the dense scene flow~\cite{Vogel13ICCV}
by a sparse scene flow~\cite{Lenz11IV} for initialization.
Sparse scene flow is less accurate, but
has a lower processing time by several orders of magnitude.
For egomotion estimation we use the visual odometry method by~\cite{Geiger11IV}. 

We perform track mining using a computer cluster by processing chunks of 500 frames. This way, object mining of large datasets can be processed efficiently in parallel in a matter of hours.
In particular, a dataset containing~$9$~h of video and~$521,500$ frames can be processed in 5-24 hours using 1043 computing nodes.
The total runtime depends on the tracking parameters and the number of proposals per frame used for tracking. In our experiments, we input the top-$100$ proposals to the tracker in each video frame.

\subsection{Object Discovery via Clustering}
\label{subsec:category-discovery}
After running the tracker%
, we obtain a reduced set of %
 object tracks, %
each of which is either classified as one of the COCO~\cite{Lin14ECCV} categories or marked as \categoryname{unknown}.
We aim to find patterns using the \categoryname{unknown} set of tracks via clustering.
This is a challenging problem: i) we are dealing with large amounts of data, ii) the mined tracks will always contain outliers and occasionally imprecise localization of objects and iii) novel objects appear rarely (\ie they appear only in the long tail of the category distribution, see Sec.~\ref{subsec:track-dataset}).

We consider several possibilities of how to tackle this problem. Many clustering methods work in two steps. First a suitable feature representation is generated, and then the clustering is performed using these features by one of the standard clustering algorithms, such as KMeans or DBSCAN. Recently, clustering has also been tackled in an end-to-end fashion using deep learning~\cite{Hsu16ARXIV, Hickinson17NIPSW}. In the following, we will describe the methods which we utilized for either extracting features or directly performing clustering.

\PAR{Extracting Features from a Pre-trained Network.} A simple %
method is to utilize a pre-trained network to extract features from its internal activations and optionally reduce their dimensionality. Since our aim is to cluster small crops of objects, a pre-trained object detector is well suited.

\PAR{Learning an Embedding on a Labeled Dataset.} Another possibility to obtain features is to make use of the recent advances in the area of feature embedding learning~\cite{Weinberger09JMLR, Schroff15CVPR, Song17CVPR}. The idea here is to use a labeled dataset to learn a feature embedding in which images of the same class have a small distance and images of different classes are far away.

A weakness of both approaches is that the source domain, on which the network or embedding is trained might differ from the target domain. Alternatively, one could pre-train an embedding on a different domain and iterate between clustering and re-learning the embedding on the target domain using the obtained clusterings~\cite{Yang16CVPR}. However, such an approach is slow and 
does not scale well to large amounts of data. %

\PAR{Clustering Algorithm.}
To perform clustering using a feature embedding, we propose to use the recent, hierarchical density-based clustering algorithm HDBSCAN~\cite{Campello15TKDD} due to its scalability to large datasets and its inherent ability to deal with outliers in the data. We show in Sec.~\ref{subsec:clustering-eval} that this approach outperforms simpler alternatives. As a distance measure between tracked objects, we use the Euclidean distance in the learned embedding space.

\PAR{Track Similarity Measure.} One of the central questions in clustering is how to define a distance measure between data points, in our case, object tracks. Object tracks are defined by a collection of image crops, representing the appearance of the tracked object over time. When applying the embedding network on tracks, we first extract a representative embedding vector for each track. We take the embedding vector of the crop that is closest to the mean of the embedding vectors of the track`s image crops. This proved to be more robust than simply taking the mean. After clustering, the resulting cluster label is transferred to the whole track.

\PAR{End-to-end Clustering.} Recently, Hsu \etal~\cite{Hsu16ARXIV} proposed ClusterNet, a scalable end-to-end clustering solution based on deep learning. They propose to train a Similarity Prediction Network (SPN), that is used to produce binary labels (same / different category) for image pairs. 
These labels are then used to train the actual clustering network (ClusterNet) on the unlabeled target data using a softmax output layer with a fixed number of classes corresponding to cluster labels. We perform an evaluation of ClusterNet trained on our data.

\section{Experimental Evaluation}

\begin{table}[t]
\scriptsize
  \begin{center}
    \begin{tabular}{lrr}
       \toprule
       & \textbf{KTC} & \textbf{OTC}\\ %
		\toprule
		Frames & 42,407 & 521,500 \\
		Duration (h) & 1.18 & 9.06\\
		Proposals (total) & 4,240,700 & 52,150,000\\
		Tracks (total) & 8,005 & 359,503\\
		Tracks (labeled) & 8,005 & 12,308\\
		Tracks (unknown) & 1,190 & 4,198\\
		Tracking Errors & 745 & 787\\
		\bottomrule
    	\end{tabular}
    \caption{Statistics of track mining from unlabeled videos. We achieve a significant reduction of the proposals using tracking.}%
	\label{tab:tracks-dataset}
  \end{center}
  \vspace{-10pt}
\end{table}

The KITTI Raw~\cite{Geiger13IJRR} dataset was recorded in street scenes from a moving vehicle in Karlsruhe, Germany. For our experiments we only use the stereo cameras and a subset of 1.18~h ($42,\!407$ frames) of video data.
The Oxford RobotCar dataset~\cite{Maddern17IJRR} has a similar setup as KITTI and it has been collected from a mobile vehicle in street scenes, mainly in the inner city of Oxford, UK. In our experiments, we only used the stereo setup.
In total 1,000 km have been recorded over 1 year, from which we use a representative subset of 9~h of video ($521,\!500$ frames).

\subsection{Video-Object Mining}
\label{subsec:track-dataset}
\begin{figure}[t]
\centering
\includegraphics[width=0.9\linewidth]{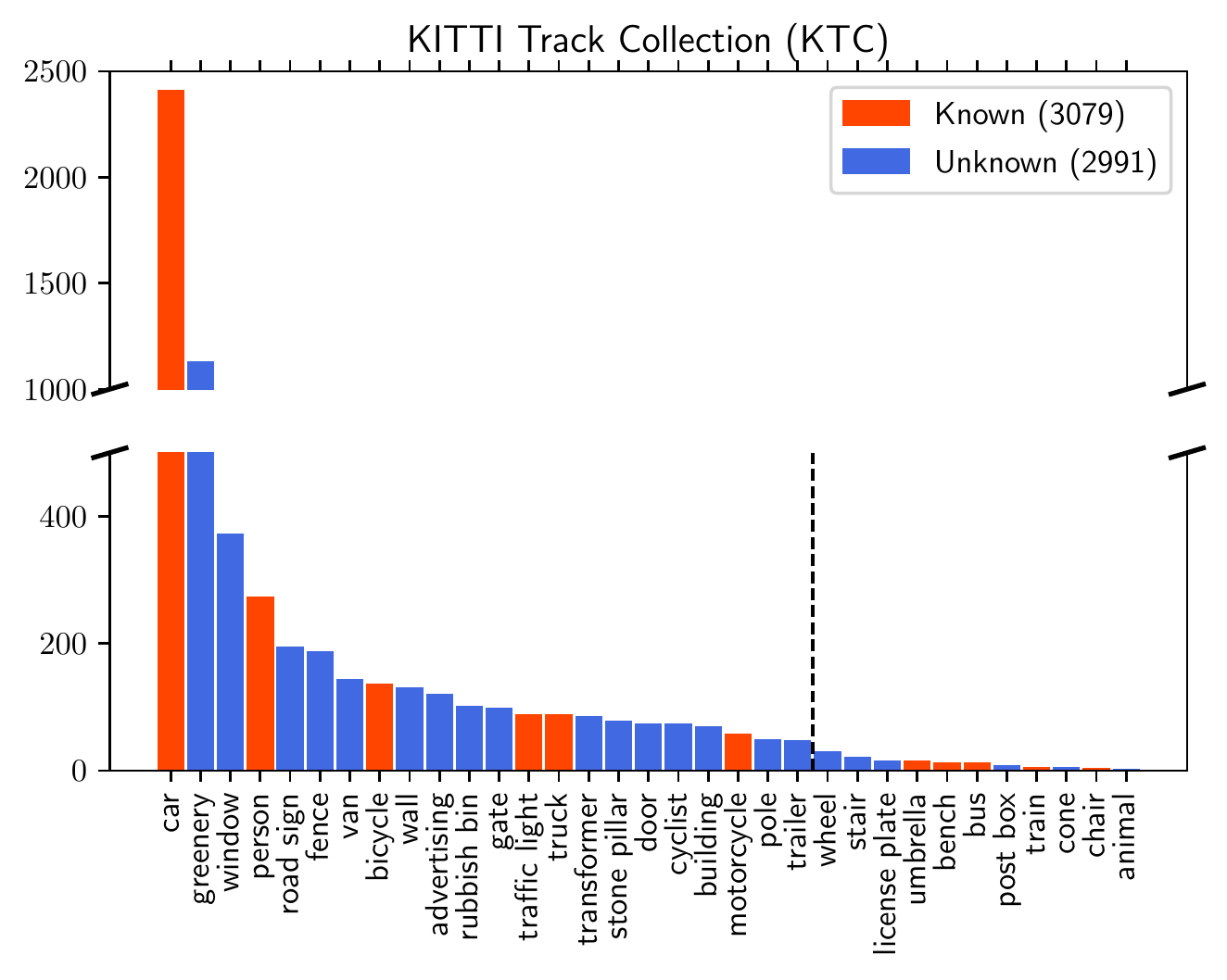}\\
\includegraphics[width=0.9\linewidth]{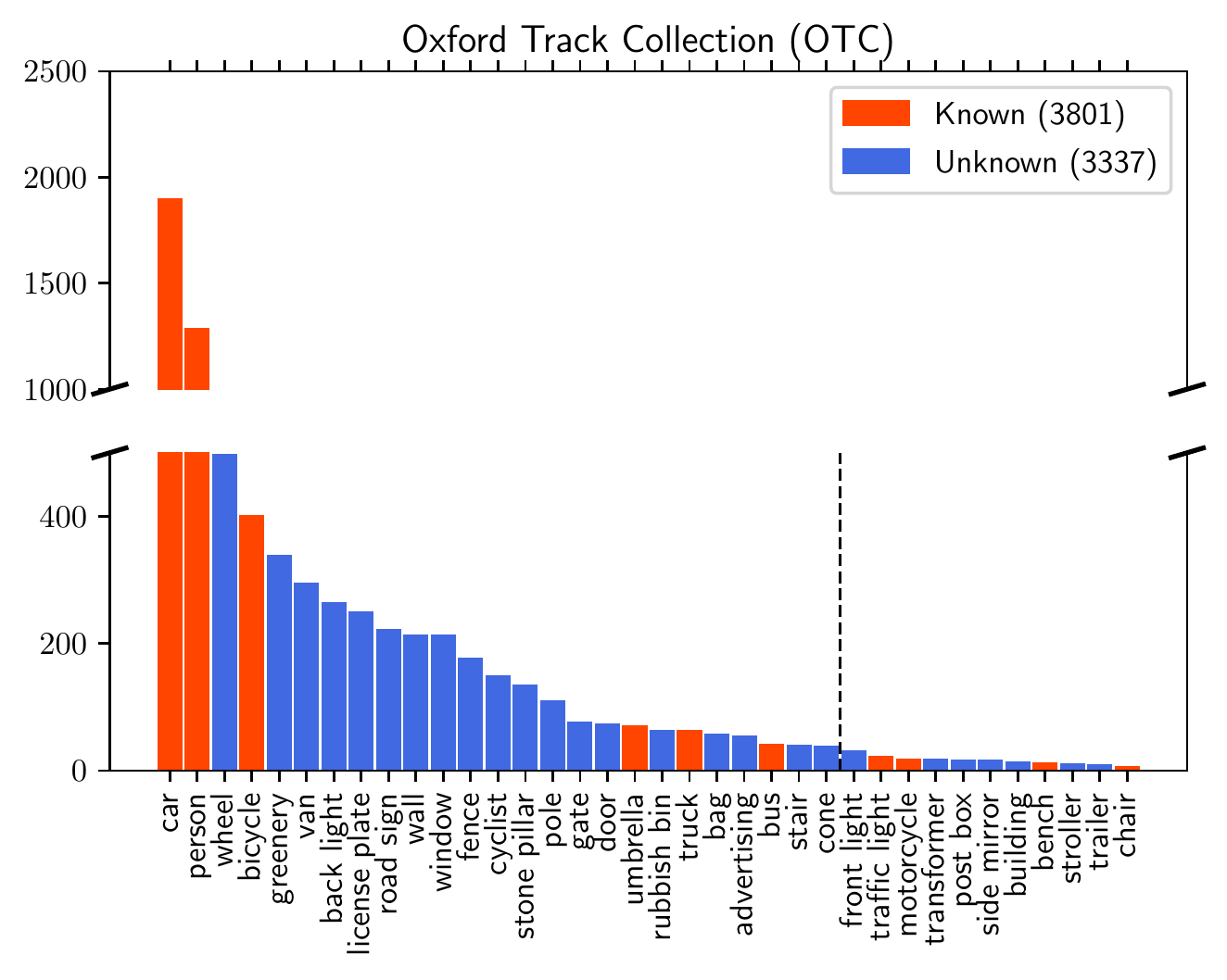}
\vspace{-8pt}
\caption{Object category distributions in KITTI Track Collection \textit{(top)} and Oxford Track Collection \textit{(bottom)}. As can be seen, the majority of the object categories appear in the long tail of the category distribution, rendering discovery of novel objects a challenging problem. The dashed line marks a cutoff, object categories beyond that are extremely rare (less than 30 instances) and are therefore excluded from the object discovery evaluation.}
\label{fig:long_tail}
\end{figure}

\begin{figure*}[ht]
    \setlength{\fboxsep}{0.3pt}%
    \setlength{\fboxrule}{0.3pt}%
    \fbox{\includegraphics[width=0.5\linewidth]{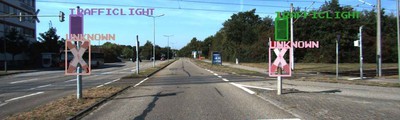}}
    \fbox{\includegraphics[width=0.50\linewidth]{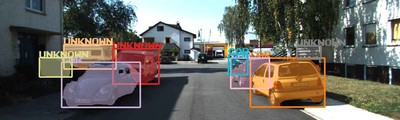}}
    \fbox{\includegraphics[width=0.50\linewidth]{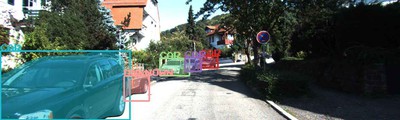}}
    \fbox{\includegraphics[width=0.50\linewidth]{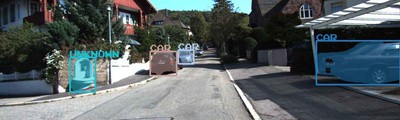}}
    \fbox{\includegraphics[width=0.50\linewidth]{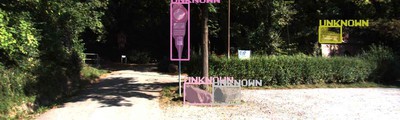}}
    \fbox{\includegraphics[width=0.50\linewidth]{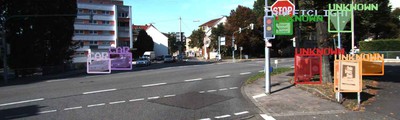}}
    \fbox{\includegraphics[width=0.50\linewidth]{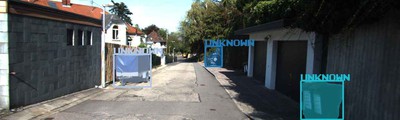}}
    \fbox{\includegraphics[width=0.50\linewidth]{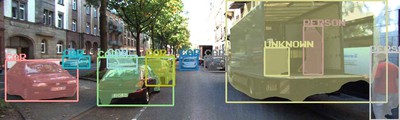}}
    \vspace{-20pt}
    \caption{Qualitative tracking results on the KITTI Raw~\cite{Geiger12CVPR} dataset. 
    Beside tracked objects, recognized by the classifier, we also find new objects such as various traffic traffic signs, car trailers, advertisements, poles, caterpillar machines, post boxes, \etc}
    \label{fig:tracking_kitti}
\end{figure*}

In this subsection, we describe and analyze the tracks we mined from the Oxford and KITTI Raw datasets. 
We input 100 mask proposals to the object tracker per frame, of which $\sim$85 pass the geometric consistency checks in a typical inner-city sequence. The tracker internally maintains on average $\sim$97 tracklet proposals per frame, of which $\sim$13 are selected as most prominent object candidates. Tab.~\ref{tab:tracks-dataset} displays a short summary of the track mining for both datasets and Fig.~\ref{fig:tracking_kitti} and Fig.~\ref{fig:tracking_oxford} show qualitative tracking results, obtained on KITTI Raw and Oxford RobotCar datasets, respectively. 
Even state-of-the-art object proposal approaches require an extremely large number of object candidates to achieve high recall for such sequences, rendering direct object discovery from proposals infeasible. Using tracking, we are able to reduce the number of object hypotheses to a manageable level and achieve a significant compression factor per image (\ie, from 100 mask proposals per image to $\sim$13 object tracks), and an even greater compression factor on the sequence level. %
\begin{figure*}[t]
    \begin{center}
        \includegraphics[width=0.24\linewidth]{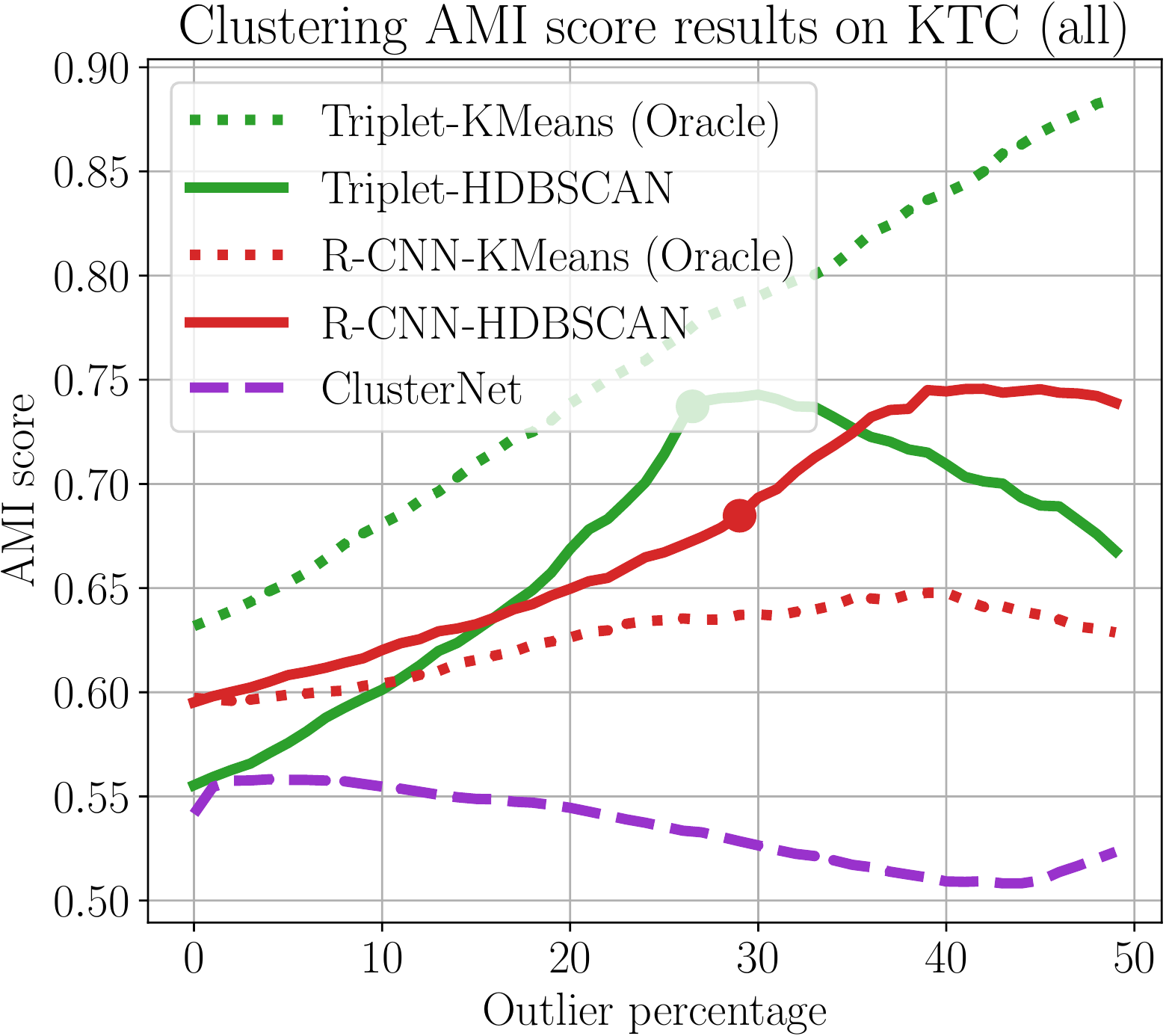}
        \includegraphics[width=0.24\linewidth]{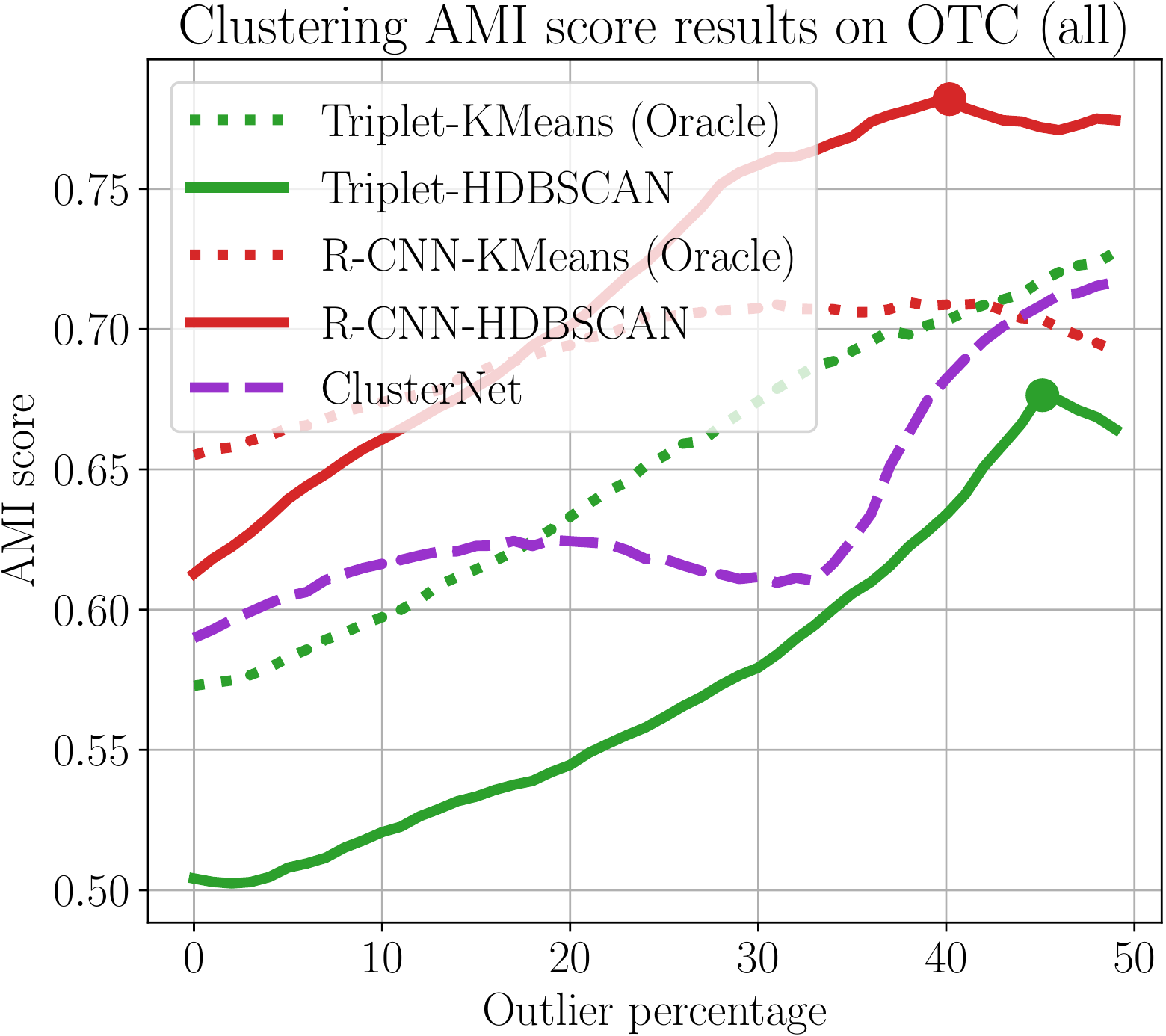}
        \includegraphics[width=0.24\linewidth]{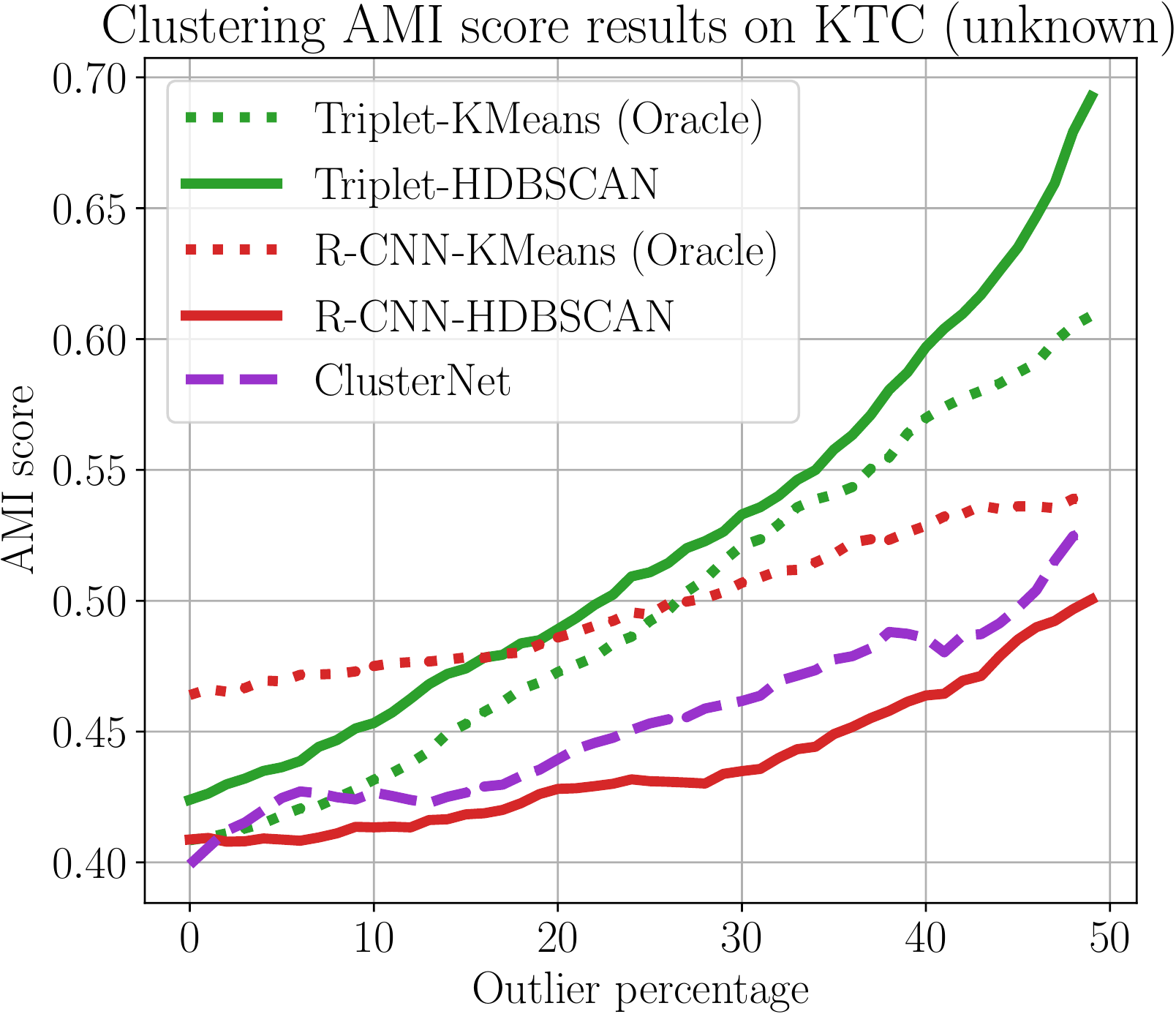}
        \includegraphics[width=0.24\linewidth]{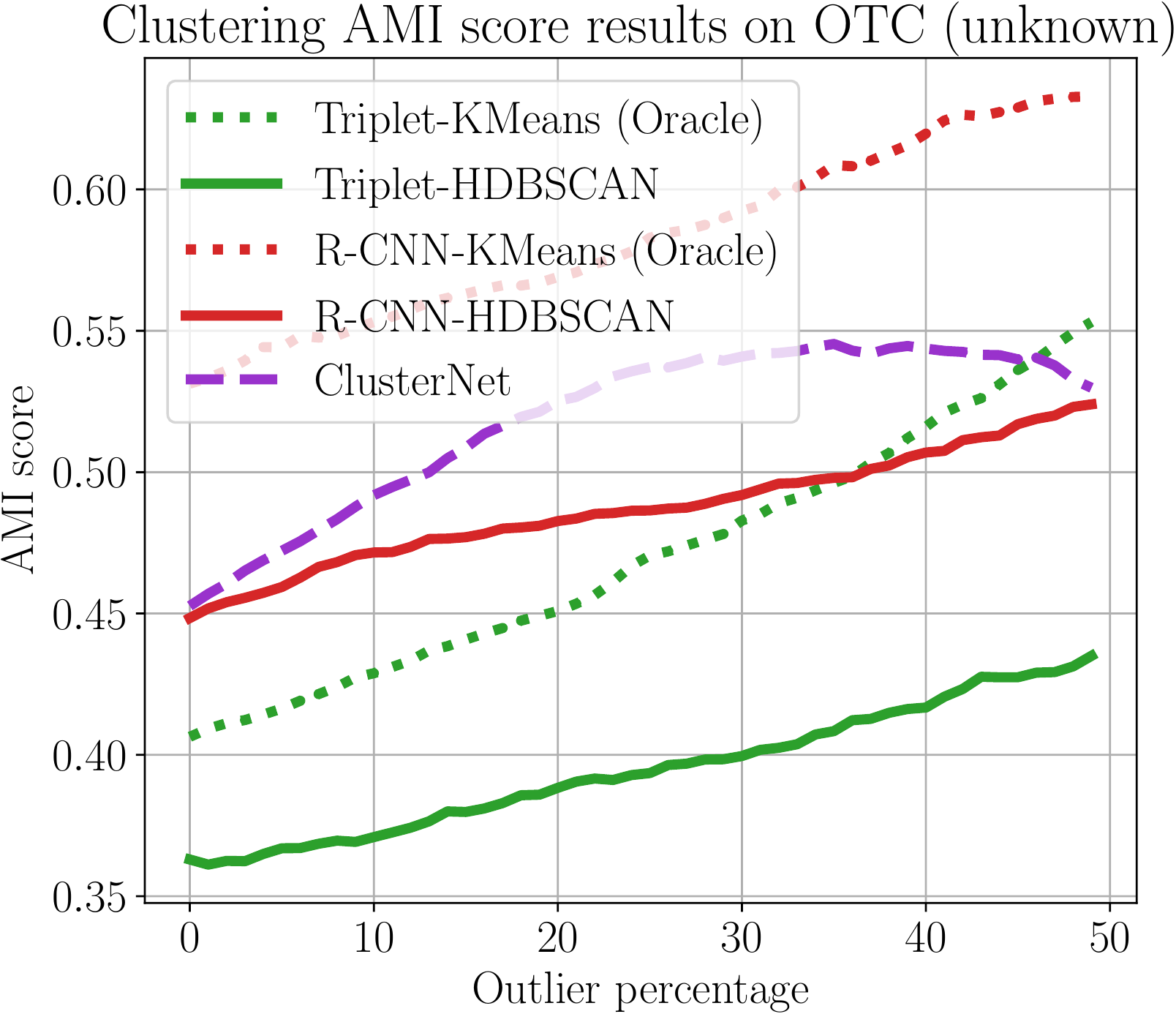}
    \end{center}
    \vspace{-15pt}
    \caption{Clustering results measured by AMI (all objects). Circle markers represent the automatically selected fraction of outliers by HDBSCAN. ``Oracle'' means that the ground truth number of clusters is used as $k$.}
\label{fig:clustering-eval-all}
\label{figurelabel}
\end{figure*}

For the purpose of a detailed analysis of tracks and clustering evaluation, we manually annotate all $8,\!005$ tracks mined on the KITTI Raw dataset and a subset of $12,\!308$ tracks, mined from Oxford RobotCar dataset. Thus, we obtain the KITTI Track Collection (KTC) and Oxford Track Collection (OTC), which we make publicly available in order to facilitate further research in the area of object discovery in automotive scenarios. These annotations have only been used for evaluation of the clusterings.
We label each track as one of 36 categories (33 in case of KTC) which we manually identified in the tracks. Tracks that diverge from the tracked object are marked as a tracking error.
When the tracked object was recognized as a valid object but does not fit into any of the 36 classes, it was labeled as a valid \categoryname{unknown} object.
Both the erroneous tracks and \categoryname{unknown} tracks are excluded for the object discovery evaluation.

As can be seen in Fig.~\ref{fig:long_tail}, the largest annotated categories in KTC are \categoryname{car}, \categoryname{greenery}, \categoryname{window}, and \categoryname{person} with $2,\!405$, $1,\!124$, $370$, and $272$ instances, respectively. A tracking error only occurred in 745 tracks ($9.3\%$) which demonstrates the robustness of the tracker. In OTC, the largest annotated categories are \categoryname{car}, \categoryname{person}, \categoryname{wheel}, and \categoryname{bicycle} with $1,\!894$, $1,\!283$, $495$, and $400$ instances, respectively. Tracking errors occur in only 787 ($2.2\%$) tracks.
Some of the categories which are not annotated in COCO are \categoryname{van}, \categoryname{trailer}, and \categoryname{rubbish bin}, for which we obtained 142, 45, and 100 instances in KTC, respectively. This demonstrates that the tracker can deliver tracks for interesting previously unseen categories, but the amount of data from the smaller KITTI Raw (1.18 h) might not yet be sufficient for discovering rare categories via clustering.

\subsection{Object Discovery}
\label{subsec:clustering-eval}
We evaluate the quality of the object discovery via clustering using the adjusted mutual information (AMI) criterion, which is a standard measure for assessing clustering performance. It measures how well the obtained clustering fits the ground truth classes.
Since the tracks contain noise, we allow the clustering algorithm to mark tracks as outliers. We then measure the performance as a function of the allowed fraction of outliers which are excluded from the evaluation. %

We compare one end-to-end trained method (ClusterNet) and a ``standard'' clustering pipeline, that utilizes trained embeddings in combination with KMeans and HDBSCAN. When running KMeans, we set the number of clusters to the ground truth number of classes to provide an upper bound on the achievable performance with KMeans. The outliers are selected based on the distance to the cluster centers.
In the following, we describe the details of each considered setup for clustering.
\subsubsection{Learned Triplet Embedding}
We train a feature embedding network on the COCO dataset~\cite{Lin14ECCV}. We apply a triplet loss~\cite{Weinberger09JMLR} to learn an embedding space with a dimensionality of $128$, in which crops of different classes are separated and crops of the same class are grouped together. To this end, we adopt the batch-hard triplet mining and the soft-plus margin formulation of~\cite{Hermans17ARXIV}. %
We trained the network to discriminate between the 80 object classes in the COCO dataset. %

\subsubsection{Last Layer of Faster R-CNN Detector} We use the activations of the last layer before the classification layer of the Faster R-CNN based detector with an Inception-ResNet-v2~\cite{Szegedy17AAAI} backbone which is also used in the tracker. For efficiency, we reduced their dimensionality from $1,536$ to $50$ using PCA, and found that the results are not very sensitive to the exact choice of dimensionality.

\subsubsection{ClusterNet} In order to assess the performance of ClusterNet~\cite{Hsu16ARXIV} on our data, we trained a Similarity Prediction Network (SPN)~\cite{Zagoruyko15CVPR} as a Siamese network with a two-class softmax.  We trained the SPN on COCO to predict whether two input crops belong to the same class. We then used the SPN output to train a ClusterNet with 50 cluster labels in the output layer on the tracks to directly predict cluster labels.

For our implementation of the SPN, ClusterNet, and for the triplet embedding network, we used a wide ResNet variant with 38 hidden layers~\cite{Wu16ARXIV} pre-trained on ImageNet~\cite{Deng09CVPR} as base architecture. The crops which either come from the COCO ground truth or from tracks, were resized bilinearly to $128\times128$ pixels before they were given into the networks.

On KTC, all $8,\!005$ tracks which we use for clustering are labeled by us. On OTC, the clustering is performed on $359,\!503$ tracks, and the evaluation is done on the $12,\!308$ tracks which we labeled. For evaluation, the tracks labeled as \categoryname{unknown}, tracking error, or with less than $30$ labeled instances are excluded.
Fig.~\ref{fig:clustering-eval-all} shows the quantitative results of the clustering evaluation on KTC and OTC. We provide a separate evaluation for  i) considering all annotated ground truth categories (Fig.~\ref{fig:clustering-eval-all} \textit{left}), and ii) only for the categories which are not in COCO (Fig.~\ref{fig:clustering-eval-all} \textit{right}).
Table~\ref{tab:res_at_0} shows the results of each of our methods evaluated at zero outlier percentage for both all objects and unknown objects not in the COCO training data.
Note that for KMeans we always set K to correspond to the ground truth number of categories while HDBSCAN estimates the number of clusters automatically. 
For the task of object discovery, it is important to evaluate clustering not just on all tracks but also on the unknown tracks only, because the track collection is dominated by known categories.

\begin{figure}[t]
\begin{center}
\includegraphics[width=1.0\linewidth]{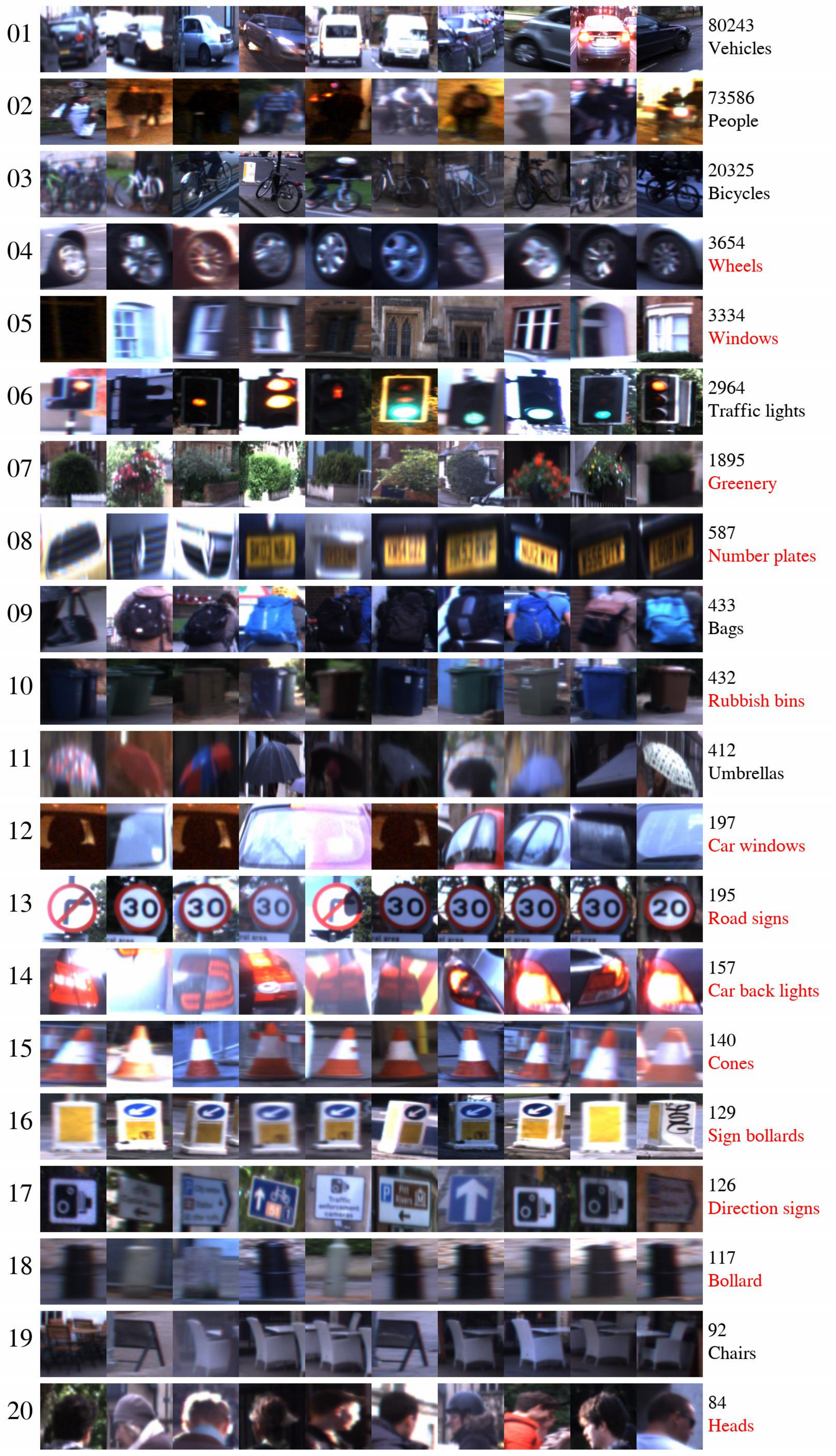}
\vspace{-20pt}
\caption{\label{fig:oxford-clusters}
Visualization of the clustering results on OTC using R-CNN features and HDBSCAN. The numbers on the right hand side indicate the number of tracks in each cluster. The cluster labels were assigned by hand. Newly discovered categories are marked in red.}
\end{center}
\end{figure}

\begin{table}[t]
\scriptsize
  \begin{center}
    \begin{tabular}{lrrrr}
       \midrule
       & \multicolumn{2}{c}{\textbf{All}} & \multicolumn{2}{c}{\textbf{Unknown}} \\
       & KTC & OTC & KTC & OTC \\ 
		\midrule   
       
		Triplet-KMeans (Oracle) & \textbf{0.63} & 0.58          & 0.40          & 0.41          \\
		Triplet-HDBSCAN        & 0.55 		   & 0.51          & 0.43          & 0.36          \\
		R-CNN-KMeans (Oracle)   & 0.60 		   & \textbf{0.65} & \textbf{0.47} & \textbf{0.53} \\
		R-CNN-HDBSCAN          & 0.60 		   & 0.62          & 0.41          & 0.45          \\
		ClusterNet             & 0.54 		   & 0.59          & 0.40          & 0.45          \\
		\midrule   
    	\end{tabular}
    \caption{Results of each of our methods evaluated at zero outlier percentage for both all objects and unknown objects not in the COCO training data. }
	\label{tab:res_at_0}
  \end{center}
\end{table}

\section{Discussion}
When evaluating different object discovery pipelines, we found that the last layer activations of a Fast R-CNN detector are surprisingly effective as a feature representation for clustering,
outperforming the learned embedding dataset and ClusterNet. This is not only the case when clustering \categoryname{known} object categories. These features achieve good performance also when clustering only \categoryname{unknown} tracks, which suggests that they generalize well and are very well suited for clustering tasks.
ClusterNet performs poorly when only a small amount of data is available but significantly improves when increasing the number of tracks, showing great potential for clustering of large amounts of unlabeled data.
KMeans (Oracle) is often the best-performing method, but here we use the ground truth number of categories as $k$, which is unrealistic in practice.
HDBSCAN performs very well, often on par with KMeans (Oracle).

\begin{figure}[t]
\begin{center}
\includegraphics[width=1.0\linewidth]{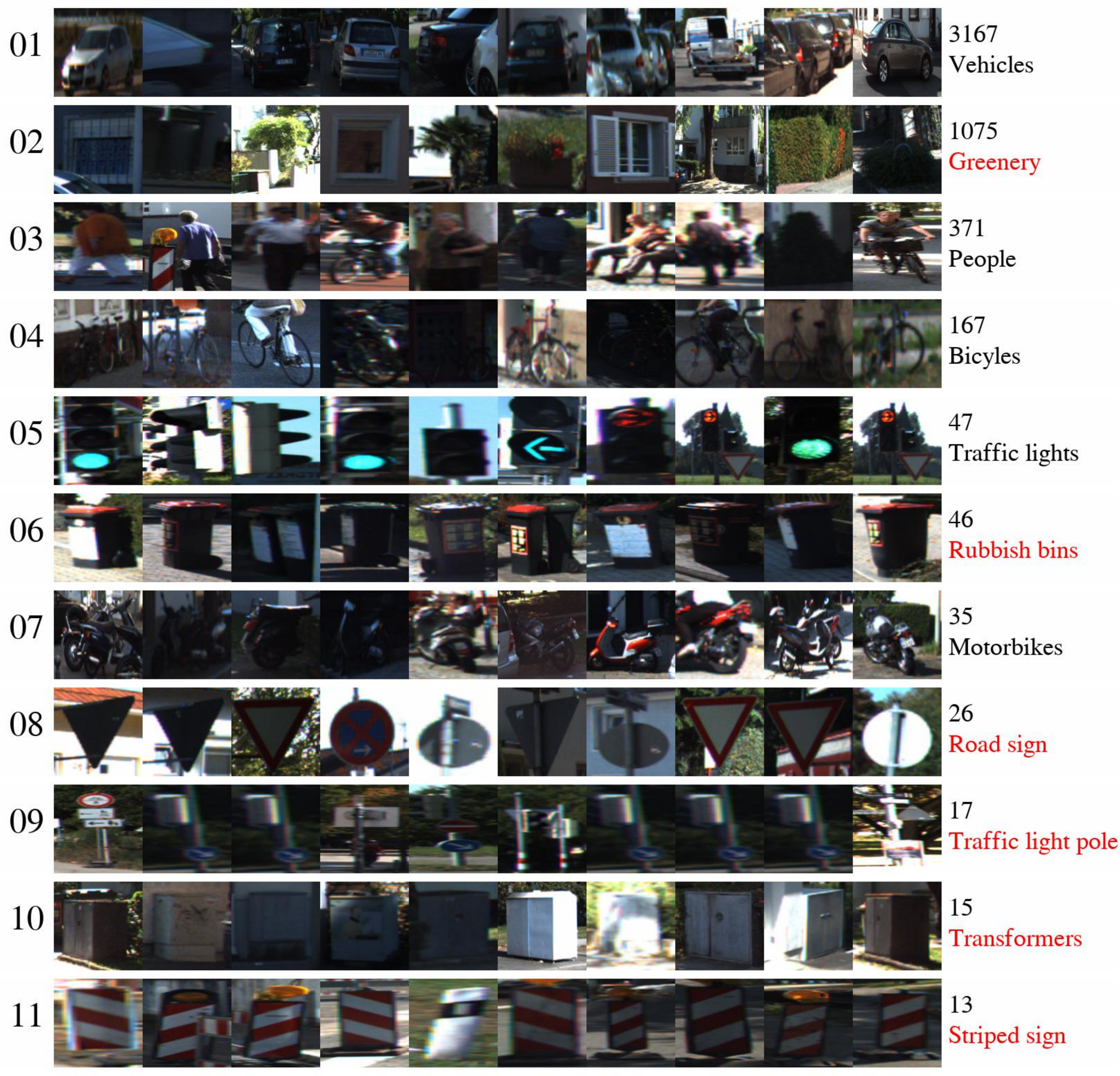}
\vspace{-20pt}
\caption{\label{fig:kitti-clusters}
Visualization of the clustering results on KTC using R-CNN features and HDBSCAN. The numbers on the right hand side indicate the number of tracks in each cluster. The cluster labels were assigned by hand. Newly discovered categories are marked in red.}
\end{center}
\end{figure}

We show qualitative results of all clusters with a size of at least 80 tracks of the obtained clustering on Oxford using R-CNN features and HDBSCAN in Fig.~\ref{fig:oxford-clusters}. As can be seen, we obtain clusters for several object types, that are not present in the COCO dataset (highlighted in red in the figure): \categoryname{wheel}, \categoryname{window}, \categoryname{greenery}, \categoryname{number plate}, \categoryname{rubbish bin}, \categoryname{car window}, \categoryname{road sign}, %
\categoryname{car back light}, \categoryname{cone}, \categoryname{sign bollard}, \categoryname{direction sign}, \categoryname{bollard}, and \categoryname{head}. On KTC (Fig.~\ref{fig:kitti-clusters}) we identify the following novel object categories: \categoryname{greenery}, \categoryname{rubbish bin}, \categoryname{road sign}, \categoryname{traffic light pole}, \categoryname{transformer}, and \categoryname{striped sign}.

\newcommand{\mysizeoxf}{0.328}
\begin{figure*}[ht]
    \setlength{\fboxsep}{0.3pt}%
    \fbox{\includegraphics[width=\mysizeoxf\linewidth]{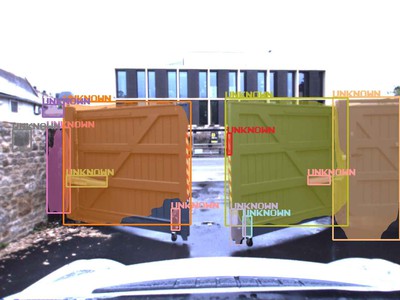}}
    \fbox{\includegraphics[width=\mysizeoxf\linewidth]{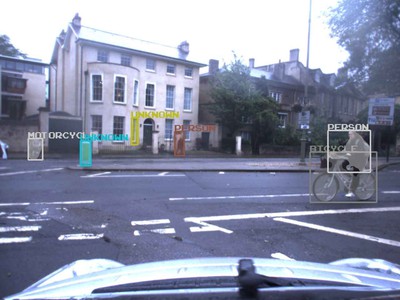}}
    \fbox{\includegraphics[width=\mysizeoxf\linewidth]{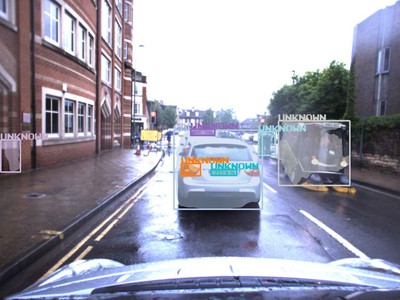}}
    \fbox{\includegraphics[width=\mysizeoxf\linewidth]{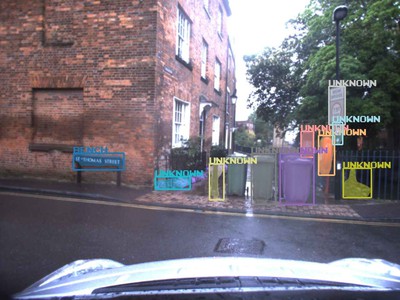}}
    \fbox{\includegraphics[width=\mysizeoxf\linewidth]{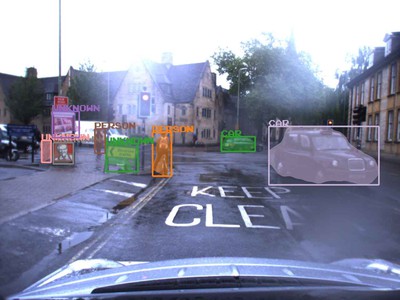}}
    \fbox{\includegraphics[width=\mysizeoxf\linewidth]{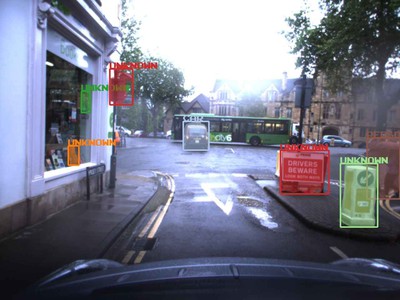}}
    \vspace{-20pt}
    \caption{Qualitative tracking results on the Oxford RobotCar~\cite{Maddern17IJRR} dataset. Beside tracked objects, recognized by the classifier, we also find new objects such as various traffic signs, traffic cones, advertisements, poles, post boxes, street cleaners, \etc}
    \label{fig:tracking_oxford}
\end{figure*}

\section{Conclusion}

This work is an initial study about object discovery from unlabeled video by automatically extracting generic object tracks. We showed that it is indeed possible to automatically discover previously unseen categories through clustering. We collected over $350,\!000$ object tracks and manually labeled over $18,\!000$ of them in order to facilitate further research in the area of object discovery.
We believe that this work is a starting point and there is still a large potential for further exploiting such unlabeled data. For example, the automatically clustered tracks could be used to fully-automatically train object detectors for the newly discovered categories.

\footnotesize \PAR{Acknowledgements:} We would like to thank Bin Huang and Michael Krause for annotation work. This project was funded, in parts, by ERC Consolidator Grant DeeVise (ERC-2017-COG-773161). The experiments were
performed with computing resources granted by RWTH Aachen University under project rwth0275.

\clearpage
{\small
\bibliographystyle{ieee}
\bibliography{abbrev_short,icra19}
}

\end{document}